# SG-FCN: A Motion and Memory-Based Deep Learning Model for Video Saliency Detection

Meijun Sun, Ziqi Zhou, Qinghua Hu, Zheng Wang, and Jianmin Jiang

*Abstract*—Data-driven saliency detection has attracted strong interest as a result of applying convolutional neural networks to the detection of eye fixations. Although a number of image-based salient object and fixation detection models have been proposed, video fixation detection still requires more exploration. Different from image analysis, motion and temporal information is a crucial factor affecting human attention when viewing video sequences. Although existing models based on local contrast and low-level features have been extensively researched, they failed to simultaneously consider interframe motion and temporal information across neighboring video frames, leading to unsatisfactory performance when handling complex scenes. To this end, we propose a novel and efficient video eye fixation detection model to improve the saliency detection performance. By simulating the memory mechanism and visual attention mechanism of human beings when watching a video, we propose a step-gained fully convolutional network by combining the memory information on the time axis with the motion information on the space axis while storing the saliency information of the current frame. The model is obtained through hierarchical training, which ensures the accuracy of the detection. Extensive experiments in comparison with 11 state-of-the-art methods are carried out, and the results show that our proposed model outperforms all 11 methods across a number of publicly available datasets.

## I. Introduction

WHEN VIEWING visual images or videos, the human visual attention mechanism helps humans selectively choose salient areas or points upon which to fixate their gaze. When observing a static image, features such as color, contour, and luminance may be dominant factors influencing the point of focus. When watching a video, however, the motion information mainly affects the human gaze. If the viewing is task driven, the human brain memory mechanism will be activated, and people will focus on a target object with high-level semantic information. In computer science, saliency detection has been widely researched in recent years to further understand and simulate the human attention mechanism. The overall efforts in this field can be divided into two main categories. The first category is salient object detection, which aims at accurately extracting objects that grab a person's attention. The second category is called eye fixation detection, which focuses on selecting a number of locations and points that may attract attention.

Images have always been the focus of computer vision research, including hyperspectral images [1], [2], and ordinary images of three channels. Saliency detection, as a preprocessing step, is an important branch in the study of images, which has received more attention over recent years and is widely used in many visual applications, including image retrieval [3], object segmentation [4], scene classification [5], object detection [6], and target tracking [7]–[11]. As for video saliency detection, to detect the significance of each frame more accurately, intraframe saliency detection needs to be carried out along with a simultaneous consideration of interframe motion and temporal information.

This paper essentially focuses on eye fixation predictions inside a video sequence. Differing from image analysis, the analysis of video sequences presents more challenges due to the fact that the motion and temporal information affects the attention of the viewer. In addition, movie videos with complex scenes and moving objects make the eye fixation detection even more difficult. Although some models based on local contrast and low-level feature information have been reported in the literature, such models often lack consideration of the interframe motion and temporal information, leading to an unsatisfactory performance when handling complex scenes, such as those with fast moving objects or a moving lens. In contrast with existing methods, we consider the motion and memory information simultaneously with the spatial information. As shown in Fig. 1, our proposed model primarily uses the designed step gained fully convolutional network with expanded information [model SGF(E)] for video fixation detection. Our model takes the saliency predictions in previous frame, the moving object boundary map between two adjacent frames, and the current frame as the input, and computes the spatiotemporal saliency probability to produce a saliency detection output, without requiring any preprocessing. Some sample predictions are given in Fig. 2.

Manuscript received January 2, 2018; revised March 9, 2018 and April 9, 2018; accepted April 19, 2018. This work was supported by the National Natural Science Foundation of China under Grant 61572351, Grant 61772360, Grant 61732011, and Grant 61620106008. This paper was recommended by Associate Editor H. Lu. *(Corresponding authors: Zheng Wang; Jianmin Jiang.)*

M. Sun, Z. Zhou, and Q. Hu are with the School of Computer Science, Tianjin University, Tianjin 300350, China (e-mail: sunmeijun@tju.edu.cn; ziqizhou@tju.edu.cn; huqinghua@tju.edu.cn).

Z. Wang is with the School of Software Engineering, Tianjin University, Tianjin 300350, China (e-mail: wzheng@tju.edu.cn).

J. Jiang is with the Research Institute for Future Media Computing, College of Computer Science and Software Engineering, Shenzhen University, Shenzhen 518060, China (e-mail: jianmin.jiang@szu.edu.cn).

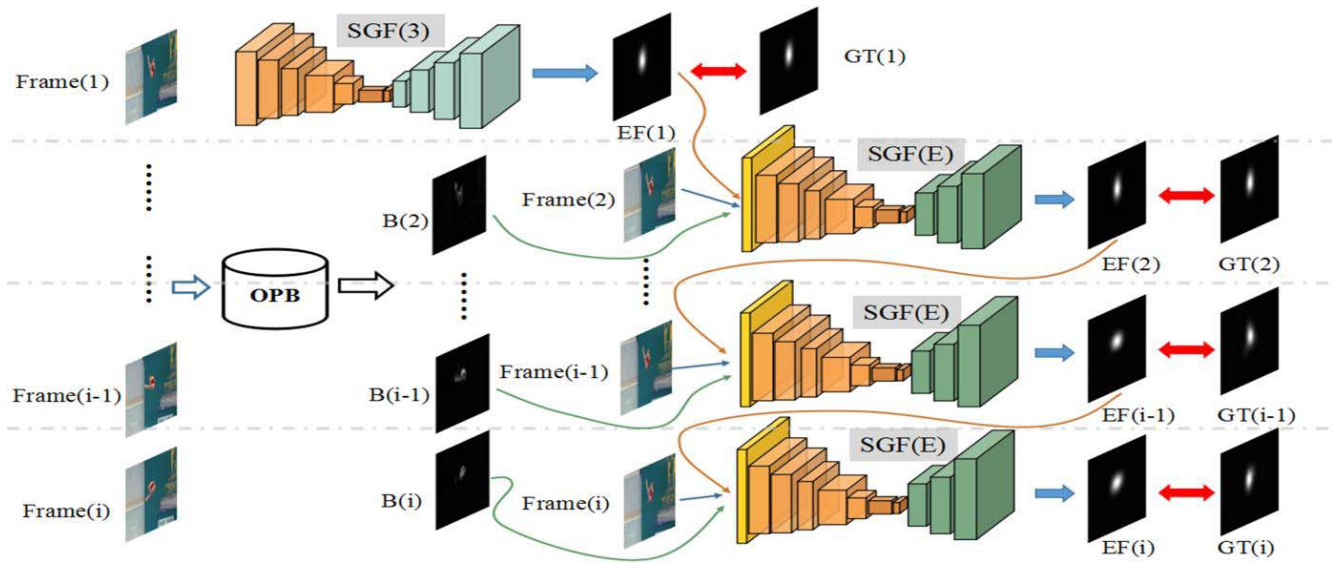

Fig. 1. Flow chart of our proposed model, in which we use the proposed model SGF for capturing the spatial and temporal information simultaneously. SGF(3) is used to handle the first frame because neither motion nor temporal information is available. From the next frame onward, the SGF(E) model takes EF(1) from SGF(3), a fast moving object edge map B(2) from the OPB algorithm, and the current frame (2) as the input, and directly outputs the spatiotemporal prediction EF(2). Section IV provides further details regarding the proposed model.

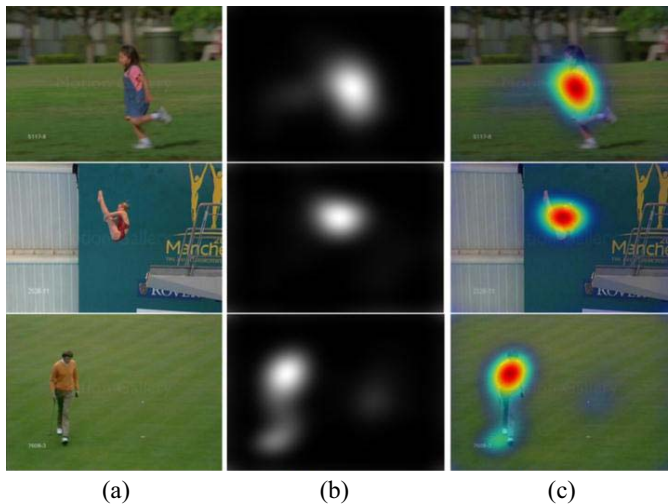

Fig. 2. Detection results from our proposed model SGF(E). (a) Raw frame. (b) Eye fixation prediction map of SGF(E). (c) Heatmap obtained from (b).

Our contributions are highlighted as follows.
1) We propose a novel and efficient model SGF, which takes the current frame, the saliency maps in previous frame, and the moving object boundary map as the input, and outputs a spatiotemporal prediction that ensures the time and space consistency.
2) We design an object palpably boundary-map (method OPB) algorithm, the details of which are shown in Section IV-C to calculate the contours of the most distinctive moving object as auxiliary motion information. The specific process is shown in Fig. 1.
3) We directly detect the eye fixation locations through the deep model without any preprocessing, which illustrates the robustness and efficiency of our model.

The rest of this paper is organized as follows. Section II introduces related works, Section III describes our method for computing the ground truth, and Section IV demonstrates the proposed model structure. The training details of our model are given in Section V, and Section VI reports the quantitative and qualitative experimental results. Finally, we provide some concluding remarks in Section VII.

## II. RELATED WORKS

Over the past few years, a number of methods have been designed for saliency detection, including traditional methods [12]–[30] and deep-learning-based methods [31]–[44]. We only introduce those works that are related to this paper.

### A. Eye Fixation Detection

Most existing fixation prediction methods are motivated by the idea that locations distinctive from the surrounding areas will attract greater attention. Thus, they have mainly aimed at determining the uniqueness and distinctiveness at a given location. To obtain the final prediction, two techniques are primarily carried out: 1) feature extraction and 2) contrast comparison.

For early feature extraction, Itti *et al.* [12] first proposed a highly influential model that considers three low-level features: 1) color; 2) orientation; and 3) intensity. Judd *et al.* [13] utilized other patterns such as steerable pyramid filters and histograms to extract low-level features including a 3-D color histogram and local energy. Lin *et al.* [14] proposed a computational visual-attention model for static and dynamic saliency maps. The Earth mover's distance (EMD) is used to measure the center-surround difference in the receptive field and to combine different features using biologically inspired nonlinear operations. Harel *et al.* [15] applied a graph-based visual saliency model, which first form activation maps on certain feature channels, and then normalize them by highlighting the conspicuity. Seo and Milanfar [16] employed a matrix cosine similarity to compute the local regression kernels and measure the likeness of a pixel to its surroundings. Guo et al. [17] believed that the phase spectrum of the Fourier transform obtains the location of salient areas, and Wang et al. [18] consider both the motion and appearance information through a quaternion feature representation. Other methods such as sparse coding [19], [20] have also been applied in certain models. However, all of the above methods use handcrafted features, which require domain-specific knowledge. Some deep-learning-based methods have recently been proposed for more accurate and simpler feature extraction, including autoencoders [43], convolutional neural network [35]–[41], and long short-term memory [44]. For example, Han et al. [43] developed a stacked denoising autoencoder to represent features from raw images. Wang et al. [35] used predetection results to enhance the estimation through a recurrent fully convolutional network.

After feature extraction, a significant step is to integrate the contrast for obtaining the final prediction. Lin et al. [14] computed the final saliency by using the Lm-norm and combine the super features using the Winner-Take-All mechanism. Wang et al. [18] used an inverse quaternion Fourier transform to reconstruct the final abnormal saliency map, Tavakoli et al. [31] proposed a framework based on interimage similarities and an ensemble of extreme learning machines. In a study by Li et al. [41], set up a multitask learning scheme for exploring the intrinsic correlations between saliency detection and semantic image segmentation and use a graph Laplacian regularized nonlinear regression model for saliency refinement. Tavakoli and Laaksonen [42] employed independent subspace analysis to obtain a hierarchical representation and exploit local and global saliency concept to achieve salient detection. Methods such as [25]–[28] utilize the fact that multiple images with common foreground can be detected simultaneously. Feature extraction and integration are performed within the range of multiple images, which leads to the problem of co-saliency detection. In addition to the above methods, using end-to-end models [35]–[44] directly to produce the prediction has remained an interesting research direction in recent years.

### B. Video Saliency Detection

According to the input, saliency models can be further categorized into static and dynamic saliency models. A dynamic saliency model takes video sequences or continual frames as an input to obtain a patch of saliency detection results. This task is also known as video saliency detection, which has recently attracted a significant interest. The reason behind this growth in popularity is the importance of video saliency detection as a preprocessing for many different tasks including video compression and summarization.

The existing video saliency models can be further divided into salient object detection models [48]–[50] and eye fixation prediction models [15], [32]–[34], [51]. The model proposed in [48] uses intraframe boundary information and interframe motion information, also combines gradient flow field with energy optimization, to achieve the spatiotemporal consistency of the output saliency maps. Wang et al. [49] introduced an unsupervised and geodesic distance-based salient video object segmentation method. They consider the spatial edges and temporal motion boundaries as indicators of foreground object locations in order to attain both spatially and temporally coherent object segmentation. Zhou et al. [50] showed a temporal filter to enhance the rendering of salient motion. Zhang and Sclaroff [51] proposed a Boolean map-based saliency model, computing the frame feature as a set of binary images and obtaining saliency maps by analyzing the topological structure of the Boolean maps. Zhong et al. [32] believed that the flow information is quite important in video detection tasks, and thus propose a fast flow model, constructing spatial and temporal saliency maps and fusing them together to create the final attention. Li and Li [33] mainly focused on compressed domain video eye fixation detection, and present an algorithm based on residual DCT coefficients norm and operational block description length. The fixation prediction is obtained through a Gaussian model whose center is determined based on the feature values. Harel et al. [15] took the motion and flicker information into account and compute the final eye fixation by adding extra channels, and Han et al. [34] built two models, a spatial attention model for predicting locations in a frame, and a temporal attention model that measures the most important frame in a video sequence.

Unlike the models mentioned above, our proposed model does not separate the spatial information detection from the temporal information detection and then fuses them. Instead, we calculate the spatial–temporal information simultaneously through a step-gained FCN (SGF) model, which is inspired from the correlation of neighboring frames inside the video sequence. In addition, we take the detection result of the previous frame as the auxiliary information, and then compute the boundaries of the moving object by exploiting the flow gradient of the adjacent frames. In this way, significant advantages can be achieved in which the information on the current frame, the saliency maps in previous frame, and the moving object boundary map are considered simultaneously to ensure the consistency in both time and space.

## III. Ground Truth Computation

In this section, we introduce how the corresponding ground truth is obtained for a video frame being observed, which is crucial for our proposed model.

### A. Dataset Introduction

For model training and performance evaluations, the ground truth fixation maps for raw videos are required. Owing to the fact that a high level of correlation exists between the human visual attention area and eye movements, we can record the eye movement data for multiple subjects using an eye tracker, and hence calculate the ground truth according to certain existing algorithms. Some representative algorithms have been reported in [34]. To implement a better training process, we compute the ground truth using two datasets: 1) the Hollywood2 dataset [45] and 2) the UCF-sports dataset [52].

For effective training, we divide these two dataset into the training set and the test set. As to the Hollywood2 dataset, the training set contains 62 000 samples and the test set contains 4500 samples. The UCF dataset consists of 21 600 training samples and 2300 testing samples.

### B. Ground Truth Computation

Similar to the work reported in [34], we proposed to calculating the ground truth using the following method. For a given video, assume there are $S$ subjects, each of which has a total of $I$ eye fixation tracking records per frame, where the total number of videos is $V$. Specifically, the ground truth value of the Hollywood2 and UCF-sports datasets can be calculated through the following three steps:

$$\left(\overline{x^j_{S_i}}, \overline{y^j_{S_i}}, k\right) = \left(\frac{\text{VR}_x(j)}{\text{SR}_x} x_{S_i}, \frac{\text{VR}_x(j)}{\text{SR}_x} \left(y_{S_i} - \frac{\text{SR}_y - \frac{\text{VR}_y(j) \cdot \text{SR}_x}{\text{VR}_x(j)}}{2}\right)\right. \\ \left. \frac{\text{currT}}{10^6} \text{fps}(j)\right). \quad (1)$$

Through (1), we obtain the true fixations from $S$ subjects per frame, where $S_i$ represents the $i$th subject. In addition, $x^j_{S_i}$ and $y^j_{S_i}$ represent the eye location coordinates for the $i$th subject of the $j$th video, respectively, and $k$ represents the frame number of the $j$th video addressed by the current coordinates. Moreover, $\text{VR}_x(j)$ and $\text{VR}_y(j)$, respectively, represent the true resolution of the $j$th video, and $\text{SR}_x$ and $\text{SR}_y$ demonstrate the horizontal and vertical resolution of the display. currT indicates the time stamp of the gaze sample (microseconds), and fps($j$) is the frame rate of the current video sequence

$$\text{my}_{\text{gauss}} = \frac{\alpha \cdot \exp\left\{-\frac{\beta \cdot \left[(\overline{R}-r)^2 + (\overline{C}-c)^2\right]}{w^2}\right\}}{\pi \cdot w} \quad (2)$$

$$G^j_k = \sum_{i=1}^{S} \sum_{t=1}^{I} \text{my}_{\text{gauss}}\left(r - y^j_{S_i} : 2r - y^j_{S_i} - 1\right. \\ \left. r - x^j_{S_i} : 2r - x^j_{S_i} - 1\right). \quad (3)$$

Here, (2) indicates the proposed Gaussian model, where the value of $W$ is empirically set to 35, indicating that a gaze point is mapped to the surrounding 35 pixels on the graph. The values of $\alpha$ and $\beta$ are empirically set, $r$ and $c$ denote the horizontal and vertical resolution of the $j$th video, respectively, and $\overline{R}, \overline{C}$ are the matrices generated from $R$ and $C$, the dimensions of which are $(2r + 1, 2c + 1)$. The final fixation map $G^j_k$ for the $k$th frame inside the $j$th video sequence can be calculated through (3).

## IV. MODEL STRUCTURE

To extract both temporal and spatial information from a video sequence simultaneously, and take the human memory mechanism into account to reflect the fact that the past frame in the video sequence will form a relatively significant object in human brain, we propose an SGF for spatial–temporal eye fixation detection.

### A. Spatial Branch: SGF(i)

The spatial branch takes a single frame as the input and produces a fixation map of the same size. We employ a fully convolutional neural network to model this process. First, we employ the first five convolutional blocks of VGGNet [53], by adding deconvolutional layers, the model ensures end-to-end detection. Second, we design three different network structures, and train them individually to design the next model based on the previous one. To reflect such a feature, we call the model SGF. As the model contains three different network structures, we can obtain features of different types and scale during the previous layers, which are useful for further training. In addition, as the deconvolutional layers have different kernel sizes, the model considers not only the global information but also the local information, producing a more accurate eye fixation map.

As shown in Fig. 4, the bottom of the network is a stack of convolutional layers. To learn more global information, we build several deconvolutional layers on the top at the 16th convolutional layer. Three models have different upsampling factors. For SGF(1), the first five convolution blocks are initialized with the weights of VGGNet, which is originally trained over 1.3 million images of the ImageNet dataset [54], and the kernel size of the deconvolutional layers is set to ×19, and ×10 with a stride of 4, respectively. The SGF(2) model is designed based on the SGF(1) model, where the parameters are initialized from SGF(1). To achieve a smoother detection, we add a deconvolutional layer and modify the kernel size to ×15 with a stride of 5, ×13 with a stride of 3, and ×22 with a stride of 2. Similarly, the SGF(3) model is based on SGF(2), where the first deconvolutional layer has ×5 upsampling factors, second, third, and the last deconvolutional layer has ×9, ×10, and ×22 upsampling factors, respectively.

In convolutional blocks, each convolutional layer needs an $h1 \times w1 \times c1$ input and an output feature map with a size of $h2 \times w2 \times c2$, where $h_i, w_i, c_i (i = 1, 2)$ denote the height, width, and channel number, respectively. The first convolutional layer takes $h \times w \times 3$ raw frames as input, and produces a feature map after a linear transformation with a bias term. Assuming that each convolutional layer has a kernel with its weights set to $W$ and the offset term set to $b$, the feature map is calculated as follows:

$$x^l_j = f\left(\sum_{i \in M_j} x^{l-1}_i * w^l_{ij} + b^l_j\right) \quad (4)$$

where $M_j$ is the number of feature maps at the previous layer $l$, $x^{l-1}_i$ represents the $j$th feature map from the $(l-1)$th layer, and $f$ is a nonlinear activation function. We choose ReLU as the activation function and embed max pooling in the convolutional layers. After a convolution operation, the feature maps are sparse and down-sampled. For up-sampling, the deconvolutional layers are applied to the top of the model

$$y = U_s(f_s(\zeta, \theta_{\text{conv}}), \theta_{\text{deconv}}) \quad (5)$$

where $\zeta$ indicates the input frame data, $f_s(\cdot)$ is the convolutional operation with parameter $\theta_{\text{conv}}$, $U_s(\cdot)$ denotes the deconvolutional operation with parameter $\theta_{\text{deconv}}$, the kernel parameter $S$ is set for up-sampling. More details on their specific implementations are described in Section V.

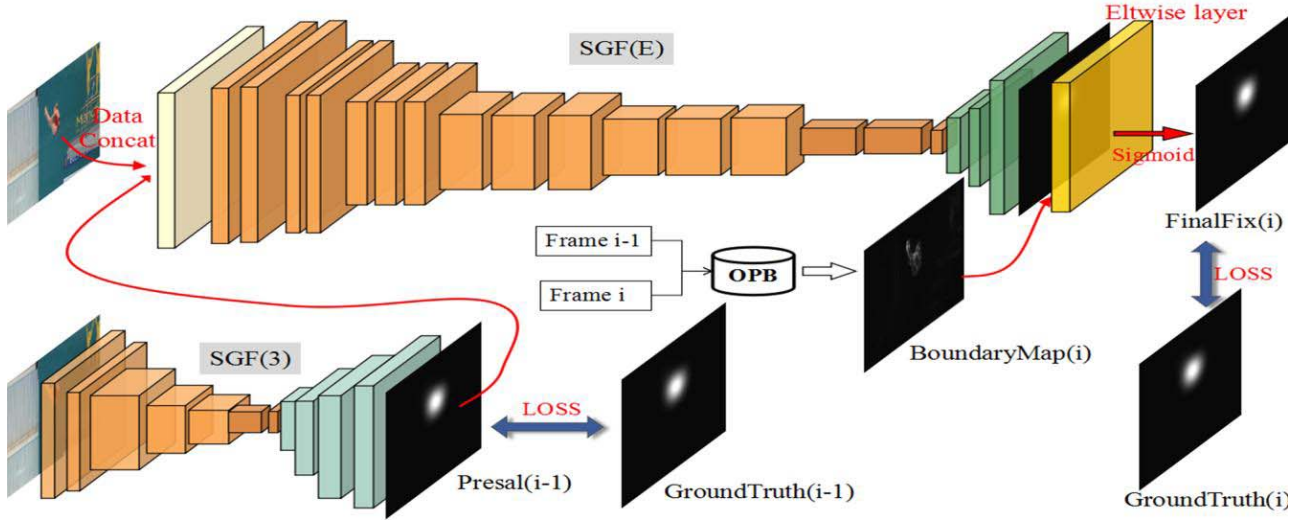

Fig. 3. Structure of model SGF(E). As shown in the flowchart, the input data is a tensor of $h \times w \times 4$. At the top of the model, we add an Eltwise layer with function MAX [big map($i$), boundary map($i$)] before Sigmoid function.

### B. Temporal Branch: SGF(E)

The temporal branch has a similar structure as the SGF(2) model as shown in Fig. 4, which takes the current frame ($i$), saliency map ($i-1$) of previous frame, and moving object boundary map ($i$) as input, and outputs the final fixation map of frame ($i$). This design is motivated by the fact that, when viewing a video sequence, not only the moving object but also fixations from the previous frame will influence the eye location in the current frame. The saliency information in previous frame added to the input and the boundary information of the moving object combined before the final detection enable and support the model to achieve a more comprehensive and accurate eye fixation identification, and such improvement is achieved only by considering the memory information regarding the previous salient object, but also by the most significant movement information of the object.

Specifically, as the first convolutional layer has an input with four dimensions, we need to concatenate the saliency map ($i-1$) and the current frame ($i$) to form the input data. Hence, the first convolutional operation is modified as

$$f(F_i, \text{pre}_{i-1}; W_{F_i}, W_{p_{i-1}}) = W_{F_i} * F_i + W_{p_{i-1}} * \text{pre}_{i-1} + b \quad (6)$$

where $W_{F_i}$ denotes the weight matrix corresponding to the input frame data $F_i$, and $W_{p_{i-1}}$ denotes the weight matrix corresponding to the saliency map $\text{pre}_{i-1}$ of previous frame. The remaining layers are set exactly like the SGF(2) model except for the last layer. Further, we use an Eltwise layer to combine the motion boundary map B($i$) with the deep map before the Sigmoid function is applied. The structure of our proposed spatiotemporal network, called SGF(E), is illustrated in Fig. 3.

**Algorithm 1** OPB for Moving Object Contour Detection

**Input:** frame $F_i$, frame $F_{i-1}$
**Output:** Boundary map $B_i$
**1:** Obtain the color gradient map $CG_i$;
  **1.1:** generate super-pixels $\{S_p^{F_i}\}$ inside the frame $F_i$ through SLIC [55];
  **1.2:** compute super-pixel segmentation map $S_i$ and the color gradient magnitude $CG_i$ using **Eq.** (8).
**2:** Obtain the optical flow gradient map $M_i$;
  **2.1:** generate the optical flow gradient magnitude $M_{F_i}$ from the optical flow map $OG_{F_i}$ through LDME [56];
  **2.2:** set a threshold $\theta$ to obtain a motion area with a higher magnitude than $\theta$.
**3:** Combine $CG_i$ and $M_i$ to obtain the boundary map $B_i$ using **Eq.** (10).

### C. OPB: Motion Information From Interframes

Our observation reveals that moving objects are more eye-catching, even though the object does not present any significant difference in comparison with the surrounding background. In other words, motion is the most crucial cue for video eye fixation detection, which makes it important for mining deeper interframe information. Following the spirit of the work reported in [48], we proposed an OPB algorithm for deep interframe information mining. As shown in Figs. 1 and 3, OPB is used to extract the contour information for the most significant moving objects through three steps, the details of which are given below.

*Step 1:* Extract the super-pixel information of the current frame to preserve the original structural elements of the video content, while effectively simplifying and ignoring some useless details. The superpixels $S^{F_i} = \{S_1^{F_i}, S_2^{F_i}, \ldots, S_p^{F_i}\}$ are distinguished by strong edges that characterize the most important content of the frame. Letting $P$ represent the number of super-pixels, where Fig. 5(b) illustrates the super-pixel segmentation map $S_i$, the color gradient magnitude at position

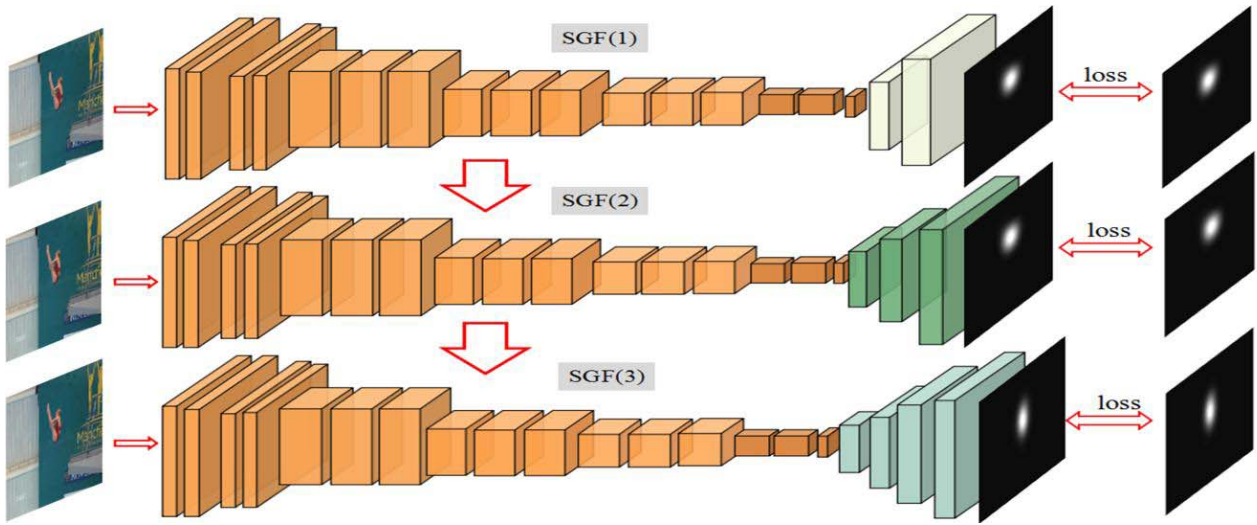

Fig. 4. Three different structures. In the SGF(1) model, convolutional blocks are initialized from VGGNet. For SGF(2) and SGF(3), convolutional blocks for the next model are initialized from the previous model. The deconvolution layers of the three models use different sizes of upsampling factors, taking into account the local and global information.

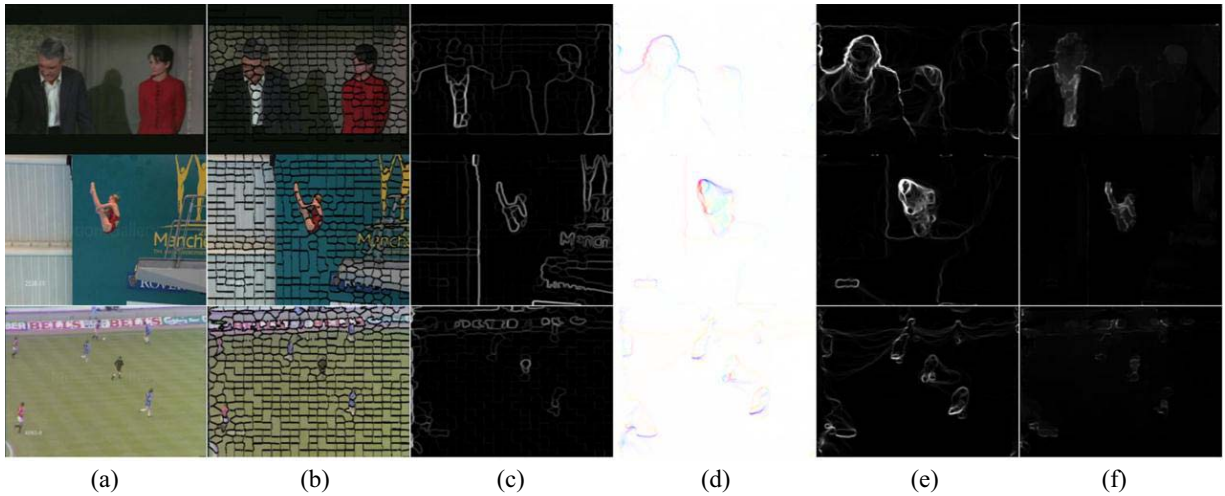

Fig. 5. Some detection results obtained through the OPB model. (a) Three different frames from different video sequences. (b) Super-pixel segmentation map $S$ using SLIC. (c) Color gradient magnitude CG from (b). (d) Visualized optical flow map $O$ of the frames in (a). (e) Optical flow gradient magnitude $M$ from (d). (f) Significant moving object boundary map $B$ achieved by fusing (c) and (e).

$z = (x, y)$ can be calculated as follows:

$$CG_i(x, y) = \|\nabla S_i(x, y)\|. \quad (7)$$

*Step 2:* Obtain an optical gradient by extracting the optical flow between the current and previous frames, and then choose those areas with a larger gradient by setting a threshold. While Fig. 5(d) illustrates some examples of such a derived optical flow, the optical gradient at position $z$ can be obtained through (8), and the magnitude $M_i(z)$ through (9)

$$\left(OG_{F_i}^x(z), OG_{F_i}^y(z)\right) = \left(\left\|\nabla O_{F_i}^z(x)\right\|, \left\|\nabla O_{F_i}^z(y)\right\|\right) \quad (8)$$

$$M_i(z) = \begin{cases} \sqrt{\left(OG_{F_i}^x(z)\right)^2 + \left(OG_{F_i}^y(z)\right)^2} & \text{if } M_i(z) > \theta \\ 0 & \text{if } M_i(z) \leq \theta \end{cases} \quad (9)$$

where $OG_{F_i}^x$ and $OG_{F_i}^y$ represent the results of the optical flow gradient along the $x$- and $y$-axes, respectively.

*Step 3:* Combine $CG_i(z)$ with $M_i(z)$ to compute the final boundary map $B_i(z)$ at position $z$ through

$$B_i(z) = \begin{cases} CG_i(z) * \left(1 - e^{-\alpha M_i(z)}\right) & \text{if } i = 1 \\ \mu B_{i-1}(z) + \lambda Pr_i & \text{if } i \geq 1 \ \& \ B_{i-1}(z) > \sigma \end{cases}$$
$$Pr_i = CG_i(z) * \left(1 - e^{-\alpha M_i(z)}\right) * \min(\|\nabla B_{i-1}(z)\|) \quad (10)$$

where $\alpha$ is a weighting factor used to decide how much boundary information of the optical gradient magnitude $M_i(z)$ need to be reserved. In our implementation, we empirically set it to 0.75. Here, $\mu$ and $\lambda$ are two scaling factors used to coordinate the calculation results. The larger $\mu$ is, the greater the influence of the previous frame. In contrast, the larger the $\lambda$, the smaller the effect of the previous frame. Here, $\sigma$ is a threshold parameter, which is also empirically set to 0.3. For the convenience of our presentation and description, a pseudo-code summary of our proposed algorithm is given in Algorithm 1, and some detection results are illustrated in Fig. 5.

## V. Training Method

In this section, we elaborate on the proposed two-stage learning to predict the human eye fixation on a video sequence. We first pretrain our model proposed in Section IV-A using the image saliency detection datasets, which enables the model to learn the features of the salient objects and grab the salient regions inside a single image. We then fine-tune the model on two video eye fixation datasets mentioned in Section III and another image eye fixation dataset reported in [57]. In this way, we ensure that the model can precisely predict the eye fixation locations. The logic behind this is that most of the benchmarks used for image saliency detection are open-sourced for public access, and the resolution of the image is relatively high with a richer variety, pretraining on these benchmarks helps to improve the model's generalization capability.

### A. Implementation Details

The SGF model is implemented based on the Caffe [58] toolbox. We initialize the first 13 convolutional layers of SGF with those of the pretrained VGG 16-layer net, and transfer the learned representations by fine-tuning to both the saliency detection and eye fixation detection. We then construct the deconvolutional layers for upsampling, the parameters of which are initialized as Gaussian distribution parameters and iteratively updated during the training. For training purposes, all images and the ground-truth maps are resized to $500 \times 500$ pixels, and the SGD learning procedure is accelerated using a NVIDIA GeForce GTX 1080ti GPU. In stage one, the momentum parameter is set to 0.99, the learning rate is set to $10^{-10}$, and the weight decay is 0.0005. In stage two, the momentum parameter is set to 0.999, the learning rate is set to $10^{-11}$, and the weight decay is 0.00005. The loss functions for two stages are designed as

$$L_1(P, G) = \frac{1}{2} \sum_{i=1}^{w} \sum_{j=1}^{h} \|G_{i,j} - P_{i,j}\|_2^2 \tag{11}$$

$$L_2(P, G) = \frac{1}{2} \sum_{i=1}^{w} \sum_{j=1}^{h} \|G_{i,j} - P_{i,j}\|_2^2 \\ + \eta \sum_{i=1}^{w \times h} [G_{i,j} \log P_{i,j} + (1 - G_{i,j}) \log(1 - P_{i,j})] \tag{12}$$

where $\eta$ is a weighting factor to show the importance of the corresponding loss item, which are set empirically along with others mentioned in earlier sections.

### B. Stage One

In this stage, we use six benchmarks related to image saliency detection. Table I shows the basic information of all six datasets intuitively. To use a cross-validation method for training, we divide all images into ten groups, and nine groups

---

**Algorithm 2** Training Method for Stage One
**Input:** image pair (I, G) for saliency detection
**Output:** pixel-wise binary map P
**1. for i=1: 3**
**2.** If i=1:
 Initialize the parameters $\omega^c_{SGF(i)}$ of the shared fully convolutional part using the pre-trained VGGNet;
**3.** Else,
 Initialize the parameters $\omega^c_{SGF(i)}$ from $\omega^c_{SGF(i-1)}$;
**4.** Initialize the parameters $\omega^d_{SGF(i)}$ of the deconvolutional part randomly from the Gaussian distribution;
**5.** Based on $\omega^c_{SGF(i)}$ and $\omega^d_{SGF(i)}$, utilize SGD and BP to train SGF(i) by minimizing the training loss using **Eq. (11)**
**6. end for**

---

**Algorithm 3** Training Method for Stage Two
**Input:** frame pair (F, G) for eye fixation detection
**Output:** pixel-wise probability map P
**1. for i=1: 2**
**2.** If i=1:
 Initialize the parameters $\varpi^c_{SGF(i)}$ from $\omega^c_{SGF(i)}$;
**3.** Else:
 Initialize the parameters $\varpi^c_{SGF(i)}$ from $\varpi^c_{SGF(i-1)}$;
**4.** Initialize the parameters $\varpi^d_{SGF(i)}$ of the deconvolutional part randomly from the Gaussian distribution;
**5.** Based on $\varpi^c_{SGF(i)}$ and $\varpi^d_{SGF(i)}$, utilize SGD and BP to train SGF(i) by minimizing the training loss using **Eq. (12)**.
**6. end for**

---

are randomly selected each time as the training set, with the remaining group applied as the test set. This design is applied to the training of methods SGF(1), SGF(2), and SGF(3), the details of which are summarized in Algorithm 2.

Where $\omega^c_{SGF(i)}$ and $\omega^d_{SGF(i)}$ represent the convolutional parameters and the deconvolutional parameters of the SGF(i) model, respectively.

### C. Stage Two

In this phase, we use the Hollywood2 and UCF datasets to train the SGF model based on the operational process completed in the first stage. The two datasets have more than 80 000 frames altogether from different video sequences and the corresponding ground truth map. Similar to stage one, we train three models individually and use the loss function in (12) to fine-tune the parameters. The detailed training procedure is summarized in Algorithm 3.

Here, $\varpi^c_{SGF(i)}$ and $\varpi^d_{SGF(i)}$ represent the convolutional and deconvolutional parameters for the SGF(i) model, respectively.

## VI. Experimental Results

In this section, we report the experimental results of the proposed approach for video eye fixation detection. First, we describe the five datasets and evaluation metrics used in this paper. Second, we provide the experimental results to demonstrate the advantages of our approach. We compared our method with 11 existing state-of-the-art methods, and both qualitative and quantitative analyses of the experimental results are presented.

TABLE I
INFORMATION ON THE SIX IMAGE SALIENCY DATASETS

| Dataset | MSRA | THUS | THUR | DUT-OMRON | DUTS | ECSSD | Total |
|---|---|---|---|---|---|---|---|
| Size | 1000 | 10000 | 6232 | 5168 | 15572 | 1000 | 38976 |

TABLE II
INFORMATION OF THE FIVE VIDEO DATASETS

| Dataset | Video Numbers | Subject Numbers | Resolution | Total Duration | Camera | Scenes |
|---|---|---|---|---|---|---|
| VAGBA [60] | 50 | 14 | 1920*1280 | 500s | Static | Outdoor |
| CRCNS [61] | 50 | 8 | 640*480 | 2-118 s per video | Dynamic & Static | Outdoor & Indoor |
| DIEM [62] | 84 | 50 | 720*576-1280*720 | 120-180 s per video | Dynamic & Static | Outdoor & Indoor |
| HOLLYWOOD2 [45] | 824 | 16 | 528*224-720*528 | 20.1 h | Dynamic & Static | Movie |
| UCF [52] | 150 | 16 | 480*360-720*576 | 958 s | Dynamic & Static | Sports |

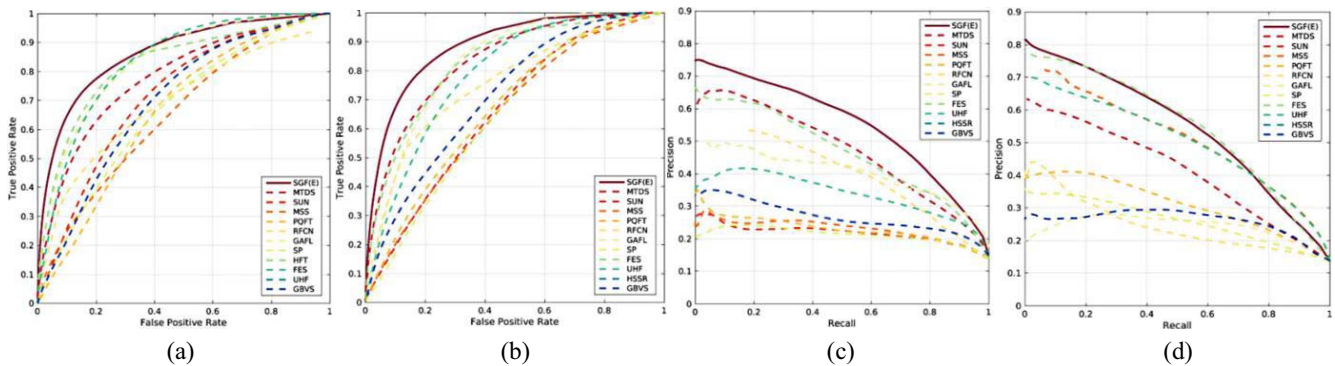

Fig. 6. Precision-recall and ROC curves gained from 12 models. (a) and (c) Hollywood2 dataset. (b) and (d) UCF-sports dataset. We can clearly see that our proposed model achieves an advanced performance compared with the others.

TABLE III
MODEL PERFORMANCE COMPARISON USING FIVE DIFFERENT METRICS ON HOLLYWOOD2

| | SGF(E) | MTDS | RFCN | GAFL | SUN | MSS | PQFT | SP | FES | UHF | HSSR | GBVS |
|---|---|---|---|---|---|---|---|---|---|---|---|---|
| sAUC↑ | **0.8857** | **0.8334** | 0.7567 | **0.8304** | 0.6620 | 0.6506 | 0.6718 | 0.6500 | 0.8291 | 0.7907 | 0.7618 | 0.7168 |
| NSS↑ | **1.4189** | **1.2464** | 0.9845 | 1.0557 | 0.5071 | 0.3606 | 0.4582 | 0.3083 | **1.1936** | 0.9890 | 0.7275 | 0.9515 |
| CC↑ | **0.5855** | 0.3561 | 0.2792 | **0.2988** | 0.1366 | 0.0945 | 0.1066 | 0.0864 | **0.3807** | 0.2510 | 0.1829 | 0.2557 |
| SIM↑ | **0.4695** | **0.2639** | 0.2297 | 0.2344 | 0.1362 | 0.1414 | 0.1420 | 0.1363 | **0.2819** | 0.1700 | 0.1490 | 0.1975 |
| EMD↓ | **2.2314** | **3.0745** | 3.5861 | 3.4407 | 4.4743 | 4.1072 | 4.4180 | 4.5863 | **2.6970** | 3.8098 | 4.2837 | 3.5416 |

The best three results are shown in red, blue, and brown, respectively.

### A. Datasets

To verify the effectiveness of the proposed model, intensive experiments were conducted on five publicly available datasets in line with the majority of previous efforts, including VAGBA [60], CRCNS [61], DIEM [62], Hollywood2 [45], and UCF [52]. Descriptions of these datasets are summarized and compared in Table II.

### B. Metrics of Evaluations

To evaluate the performance effectively, we adopted seven widely used indicators for eye fixation detection, including the receiver operating characteristic (ROC) curve, the precision recall (PR) curve, area under the ROC curve (shuffled-AUC) [63], the linear correlation coefficient (CC) [64], the similarity (SIM), the EMD [47], and the normalized scanpath saliency (NSS) [59]. ROC and AUC are generated by thresholding pixels in a fixation map into binary masks with a sequence of fixed integers from 0 to 255. The precision $P$, true positive rate TPR, and false positive rate FPR are defined, respectively, as follows:

$$P = \sum_{i \in I} \frac{|P_i \cap \text{FG}_i|}{|P_i|}, \quad \text{TPR} = \sum_{i \in I} \frac{|P_i \cap \text{FG}_i|}{|\text{FG}_i|}$$

$$\text{FPR} = \sum_{i \in I} \frac{|P_i \cap \text{BG}_i|}{|\text{BG}_i|}. \quad (13)$$

### C. Performance Comparison

To validate the advantages of our proposed model, we compared it with 11 methods, including MTDS [41], RFCN [35], GAFL [48], SUN [22], MSS [21], PQFT [17], SP [12], FES [24], UHM [42], HSSR [23], and GBVS [15]. The specific experimental results are presented below.

*1) Experimental Comparison:* To verify the performance of the proposed model (SGF), we plotted the ROC and PR

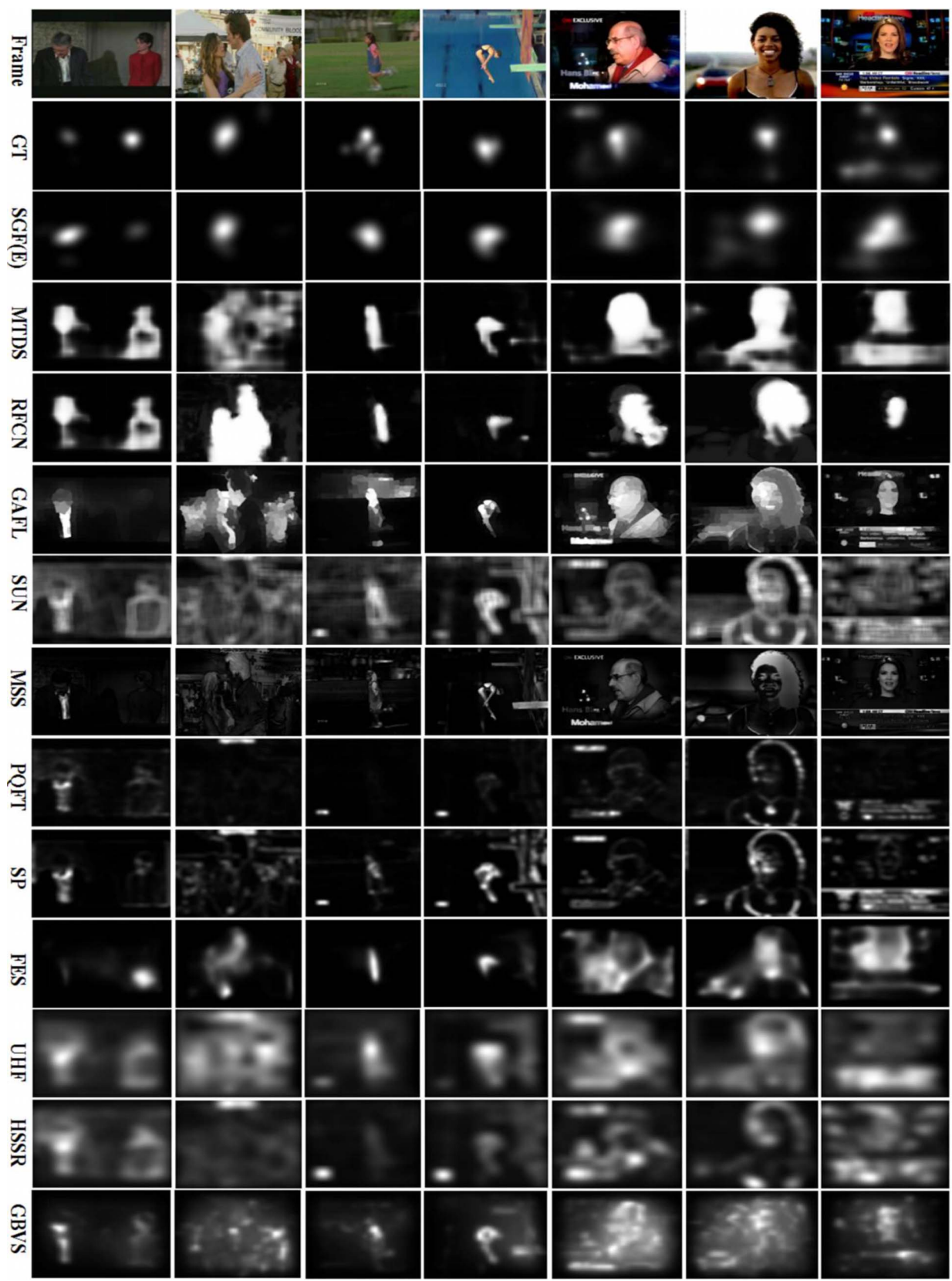

Fig. 7. Fixation prediction maps of our proposed SGF(E) model and 11 benchmarks, first two rows belong to HOLLYWOOD2, the third and fourth rows belong to UCF-sports, the fifth row belongs to VAGBA, the sixth row belongs to DIEM, and the last row belongs to CRCNS.

curves for SGF and 11 state-of-the-art methods (Fig. 6), and calculated the corresponding shuffled-AUC, NSS, SIM, and EMD of Hollywood2 and UCF datasets. Tables III and IV summarize the experimental results, and comparative experiment on VAGBA ,CRCNS, and DIEM datasets, as well as the experimental results are shown in Fig. 7. It can be seen that

TABLE IV
Model Performance Comparison Using Five Different Metrics on UCF

|  | SGF(E) | MTDS | RFCN | GAFL | SUN | MSS | PQFT | SP | FES | UHF | HSSR | GBVS |
|---|---|---|---|---|---|---|---|---|---|---|---|---|
| sAUC↑ | **0.8624** | 0.7791 | 0.6923 | **0.8350** | 0.7183 | 0.6518 | 0.6599 | 0.6669 | 0.8246 | **0.8339** | 0.6983 | 0.6839 |
| NSS↑ | **1.4493** | 1.0525 | - | 1.1908 | 0.7614 | 0.4959 | 0.3067 | 0.5378 | **1.2875** | **1.2867** | 0.6625 | 1.2035 |
| CC↑ | **0.5532** | 0.3598 | 0.3049 | **0.3881** | 0.2290 | 0.1749 | 0.1089 | 0.1979 | **0.4431** | 0.3497 | 0.1962 | 0.3604 |
| SIM↑ | **0.4555** | 0.2733 | 0.2067 | **0.3110** | 0.1585 | 0.1735 | 0.1514 | 0.1779 | **0.3577** | 0.1962 | 0.1501 | 0.2295 |
| EMD↓ | **2.4096** | 3.5393 | 3.2845 | **2.6805** | 4.4155 | 4.0080 | 4.7652 | 4.3314 | **2.2495** | 3.6832 | 4.4801 | 3.1090 |

The best three results are shown in red, blue, and brown, respectively.

TABLE V
Differences Between the Four Mentioned Models

|  | TRAINING DATASET | | MODEL STRUTRE | | | PRESAL | BOUNDARY MAP |
|---|---|---|---|---|---|---|---|
|  | SALIENT OBJECT DETECT | EYE FIX DETECT | TWO DECONV LAYER | THREE DECONV LAYER | FOUR DECONV LAYER |  |  |
| SGF (1) | √ | × | √ |  |  | × | × |
| SGF (2) | √ | × |  | √ |  | × | × |
| SGF (3) | √ | √ |  |  | √ | × | × |
| OPB | × | × |  |  |  | × | √ |
| SGF$_{nb}$ | √ | √ |  | √ |  | √ | × |
| SGF(E) | √ | √ |  | √ |  | √ | √ |

TABLE VI
Ablation Analysis of the Proposed Method on Two Datasets

|  | Hollywood2 | | | | | | UCF-Sports | | | | | |
|---|---|---|---|---|---|---|---|---|---|---|---|---|
|  | SGF(1) | SGF(2) | SGF(3) | OPB | SGF$_{nb}$ | SGF(E) | SGF(1) | SGF(2) | SGF(3) | OPB | SGF$_{nb}$ | SGF(E) |
| sAUC↑ | 0.8498 | 0.8262 | 0.8776 | 0.4989 | 0.4996 | **0.8857** | 0.8141 | 0.7751 | 0.8507 | 0.5021 | 0.5017 | **0.8624** |
| SIM↑ | 0.3569 | 0.3731 | 0.4568 | 0.1192 | 0.2338 | **0.4695** | 0.3524 | 0.3588 | 0.4473 | 0.1075 | 0.1916 | **0.4555** |
| CC↑ | 0.4321 | 0.4508 | 0.5579 | 0.2010 | 0.2387 | **0.5855** | 0.4205 | 0.4418 | 0.5355 | 0.1140 | 0.1718 | **0.5532** |
| NSS↑ | 1.2592 | 1.2674 | **1.4688** | 0.5392 | 0.5392 | 1.4189 | 1.1599 | 1.1388 | 1.4304 | 0.2497 | 0.4237 | **1.4493** |
| EMD↓ | 2.8682 | 2.9315 | 2.2396 | 3.1926 | 3.1926 | **2.2314** | 3.2626 | 3.3780 | 2.4788 | 3.8843 | 3.4966 | **2.4096** |

our model shows a better generalization capability in dealing with complicated scenes, and can accurately find eye fixation points to make predictions that are closest to the ground truth.

*2) Ablation Study:* To evaluate the performance of SGF in comparison with the four models proposed in this paper, we summarize the main differences among the four models in Table V, and the results of their performance comparisons are given in Table VI.

As shown, the SGF(E) model, which combines the motion information from OPB and the memory information from the SGF(3) model, achieves the best performance on both datasets. Correspondingly, two important conclusions can be drawn: 1) the memory information from the previous frames is useful for detections in the current frame and 2) the motion information across neighboring frames plays a constructive role in improving the overall performance through a fusion of this information with current detections.

## VII. Conclusion

In this paper, we proposed a robust deep model for the detection of video eye fixations. By studying the mechanisms of human visual attention and memory, we simulate the process of viewing video sequences by human beings, and added both memory and motion information to enable the model to capture the salient points across neighboring frames. With this process, both the previous detection and the motion information were taken into account to achieve the maximum probability of eye fixations, which improve the accuracy of the detection results. Intensive experiments validated the superiority of our proposed model in comparison with 11 representative existing state-of-the art methods.

Finally, we highlight our main contributions as follows.
1) We proposed a deep model for video saliency detection without the need of any preprocessing operations.
2) The memory information was exploited to enhance the model generalization by considering the fact that changes between two adjacent frames inside a video are limited within a certain range, and hence the corresponding eye fixations should remain correlated.
3) Extensive experiments were carried out and comparative results were reported, which not only supported that our proposed model is superior in comparison with the previous methods but also validated the robustness of our proposed approaches.

Further research can be identified to focus more on human brain activities and explore in detail the mechanism of human memory, thereby achieving more accurate and robust detections of eye fixation points as well as their saliencies.

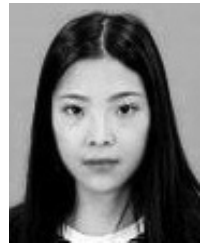

**Ziqi Zhou** is currently pursuing the M.S. degree in computer science with the School of Computer Science and Technology, Tianjin University, Tianjin, China.

Her current research interests include computer image/video processing, and particularly deep learning.

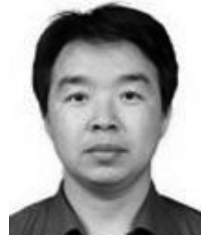

**Qinghua Hu** received the B.S., M.S., and Ph.D. degrees in computer science from the Harbin Institute of Technology, Harbin, China, in 1999, 2002, and 2008, respectively.

He was a Postdoctoral Fellow with the Department of Computing, Hong Kong Polytechnic University, Hong Kong, from 2009 to 2011. He is currently a Full Professor and the Vice Dean of the School of Computer Science and Technology, Tianjin University, Tianjin, China. He has authored over 100 journal and conference papers in the areas of granular computing-based machine learning, reasoning with uncertainty, pattern recognition, and fault diagnosis. His current research interests include rough sets, granular computing, and data mining for classification and regression.

Prof. Hu was the Program Committee Co-Chair of the International Conference on Rough Sets and Current Trends in Computing in 2010, the Chinese Rough Set and Soft Computing Society in 2012 and 2014, and the International Conference on Rough Sets and Knowledge Technology, the International Conference on Machine Learning and Cybernetics in 2014, and the General Co-Chair of IJCRS 2015. He is currently the PC-Co-Chair of CCML 2017 and CCCV 2017.

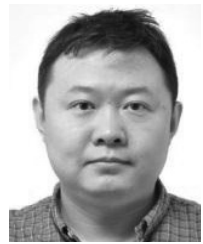

**Zheng Wang** received the Ph.D. degree in computer science from the School of Computer Science, Tianjin University (TJU), Tianjin, China, in 2009.

He is currently an Associate Professor with the School of Computer Software, TJU. He was a Visiting Scholar with INRIA Institute, Paris, France, from 2007 to 2008. His current research interests include video analysis, hyperspectral imaging, and computer graphics.

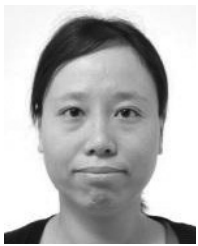

**Meijun Sun** received the Ph.D. degree in computer science from the School of Computer Science, Tianjin University (TJU), Tianjin, China, in 2009.

She is currently an Associate Professor with the School of Computer Science and Technology, TJU. She was a Visiting Scholar with INRIA Institute, Paris, France, from 2007 to 2008. Her current research interests include computer graphics, hyperspectral imaging, and image processing.

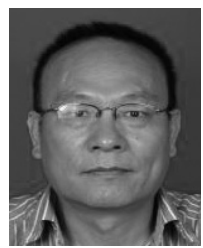

**Jianmin Jiang** received the Ph.D. degree in computer engineering from the University of Nottingham, Nottingham, U.K., in 1994.

From 1997 to 2001, he was a Full Professor of computing with the University of Glamorgan, Pontypridd, U.K. In 2002, he joined the University of Bradford, Bradford, U.K., as a Chair Professor of digital media, and the Director of the Digital Media and Systems Research Institute. He was with the University of Surrey, Guildford, U.K., as a Full Professor from 2010 to 2015 and as a Distinguished Professor (1000-plan) with Tianjin University, Tianjin, China, from 2010 to 2013. He is currently a Distinguished Professor and the Director of the Research Institute for Future Media Computing, College of Computer Science and Software Engineering, Shenzhen University, Shenzhen, China.

Dr. Jiang has been a Chartered Engineer, a fellow of IEE/IET and RSA, a member of EPSRC College in the U.K., and an EU FP-6/7 Evaluator.